\documentclass[11pt,letterpaper]{article}

\usepackage{emnlp2017}
\usepackage{times}
\usepackage{latexsym}
\usepackage{graphicx}
\usepackage{url}

\emnlpfinalcopy

\def\emnlppaperid{***}

\newcommand{\ProComp}{\textsc{ProComp}}
\newcommand{\prolocal}{\textsc{ProLocal}}
\newcommand{\proglobal}{\textsc{ProGlobal}}

\newcommand{\eat}[1]{}
\mathchardef\mhyphen="2D
\newenvironment{ite}{                     
     \parskip 0cm \begin{itemize} \parskip 0cm \parsep 0cm \itemsep 0cm \topsep 0cm}{
        \end{itemize}} 
\newenvironment{enu}{                   
     \parskip 0cm \begin{list}{}{\parsep 0cm \itemsep 0cm \topsep 0cm}}{
       \end{list}} 
\newenvironment{des}{                 
     \parskip 0cm \begin{list}{}{\parsep 0cm \itemsep 0cm \topsep 0cm}}{
       \end{list}} 

\title{What Happened? Leveraging VerbNet to Predict the Effects of Actions in Procedural Text}

\author{Peter Clark, Bhavana Dalvi, Niket Tandon \\
  Allen Institute for AI, Seattle, WA \\
  \{peterc,bhavanad,nikett\}@allenai.org}

\date{}

\begin{document}

\maketitle

\begin{abstract}
  Our goal is to answer questions about paragraphs describing processes (e.g., photosynthesis).
  Texts of this genre are challenging because the effects of actions are
often implicit (unstated), requiring background knowledge and inference
to reason about the changing world states. To supply this knowledge, we
leverage VerbNet to build a rulebase (called the Semantic Lexicon)
of the preconditions and effects of actions,
and use it along with commonsense knowledge of persistence
to answer questions about change. Our evaluation shows that
our system \ProComp~significantly outperforms two
strong reading comprehension (RC) baselines. Our contributions are two-fold: the
Semantic Lexicon rulebase itself, and a demonstration of how a simulation-based
approach to machine reading can outperform RC methods that rely on surface cues alone.

Since this work was performed, we have developed neural systems that outperform
\ProComp, described elsewhere \cite{naacl}. However, the Semantic Lexicon remains a novel
and potentially useful resource, and its integration with neural systems remains a
currently unexplored opportunity for further improvements in machine reading about
processes.
\end{abstract}

\section{Introduction}
Our goal is to answer questions about paragraphs describing processes.
This genre of texts is particularly challenging because they describe a 
changing world state, often requiring inference to answer questions about those states.
Consider the paragraph in Figure~\ref{example}.
While reading comprehension (RC) systems \cite{Seo2016BidirectionalAF,zhang2017exploring}
reliably answer lookup questions such as:
\vspace{1mm} \\
\hspace*{0mm}  (1) What do the roots absorb?(A:water,minerals) 
\vspace{1mm}   \\
they struggle when answers are not explicit, e.g.,
\vspace{1mm} \\
\hspace*{0mm}  (2) Where is sugar produced? (A:in the leaf)  
 \vspace{1mm} \\
(e.g., BiDAF \cite{Seo2016BidirectionalAF} answers ``glucose'').
This last question requires knowledge and inference: 
If carbon dioxide {\it enters} the leaf (stated), then it will be {\it at}
the leaf (unstated), and as it is then used to produce sugar, the sugar
production will be at the leaf too.
This is the kind of inference our system, \ProComp, is able to model. 

\begin{figure}
\centerline{
\fbox{%
    \parbox{0.44\textwidth}{%
	Chloroplasts in the leaf of the plant trap light from the sun.
	The roots absorb water and minerals from the soil.
	This combination of water and minerals flows from the stem into the
	leaf. Carbon dioxide enters the {\bf leaf}.
	Light, water and minerals, and the carbon dioxide all combine into a mixture.
	This mixture forms {\bf sugar} (glucose) which is what the plant eats.        
\vspace{2mm}
\begin{des}
\item[{\bf Q:}] Where is sugar produced?
\item[{\bf A:}] in the leaf
\end{des}
    }%
}}
\caption{A paragraph from \emph{ProPara} about photosynthesis (bold added, to
  highlight question and answer elements). Processes are challenging because
  questions (e.g., the one shown here) often require inference.}
\label{example}
\vspace{-5mm}
\end{figure}

\begin{table*}[tbh]
  \centering
  \small{
\begin{tabular}{|l|l|l|}\hline
Verb &   Classes   &  Rule \\ \hline
assemble  & build-26.1  & {\bf IF} (Agent ``assemble'' Product) \\
		      & & {\bf THEN before:} not exists(Product) {\bf \& after:} exists(Product) \\
assemble  & build-26.1  & {\bf IF} (Agent ``assemble'' Material ``into'' Product) \\
	              & & {\bf THEN before:} not exists(Product) {\bf \& after:} exists(Product) \\
... & & \\
enter	&  escape-51.1-2 & {\bf IF} (Theme "enter" Destination) \\
                       & & {\bf THEN before:} not is-at(Theme,Destination) {\bf \& after:} is-at(Theme,Destination) \\
enter   &  escape-51.1-2 & {\bf IF} (Theme "enter" - (PREP-src Initial\_Location)) \\
                     & &   {\bf THEN before:} not is-at(Theme,Destination) {\bf \& after:} is-at(Theme,Destination) \\
... & & \\ \hline
\end{tabular}
}
\caption{The Semantic Lexicon, a derivative and expansion of (part of) VerbNet, contains rules describing how linguistically-expressed events change the world.
  For example, for the sentence ``CO2 enters the leaf'', the first rule for ``enter'' above will fire, predicting
  that is-at(``CO2'',''leaf'') will be true after that event.}
 \label{semantic-lexicon}
\end{table*}

\eat{ **OLD VERSION**
  \begin{table*}[tbh]
{\small
\begin{tabular}{|l|l|l|l|l|}\hline
Verb &   Classes   &  Syntax & 	        			Facts true before	          &   Facts true after \\ \hline
assemble  & build-26.1  & (Agent ``assemble'' Product)  			& not exists(Product) &  exists(Product) \\
assemble  & build-26.1  & (Agent ``assemble'' Material & not exists(Product) &  exists(Product) \\
 &  & \multicolumn{1}{|r|}{("into" Product))}  & & \\
... & & & & \\
enter	&  escape-51.1-2 & (Theme "enter" Destination) & not is-at(Theme,Destination) & is-at(Theme,Destination) \\
enter   &  escape-51.1-2 & (Theme "enter" - & is-at(Theme,Initial\_Location), & not is-at(Theme,Initial\_Location) \\
... & & \multicolumn{1}{|r|}{(PREP-src Initial\_Location)} & & \\ \hline
\end{tabular}
}
\vspace{-3mm}
\caption{The Semantic Lexicon, a derivative of VerbNet, describes how events change the world. For example
in the state after (``CO2'' ``enter'' ``leaf''), is-at(``CO2'',''leaf'') will be true.}
 \label{semantic-lexicon}
  \end{table*}
  }

\begin{figure*}[bth]
\centering
\includegraphics[width=2.1\columnwidth]{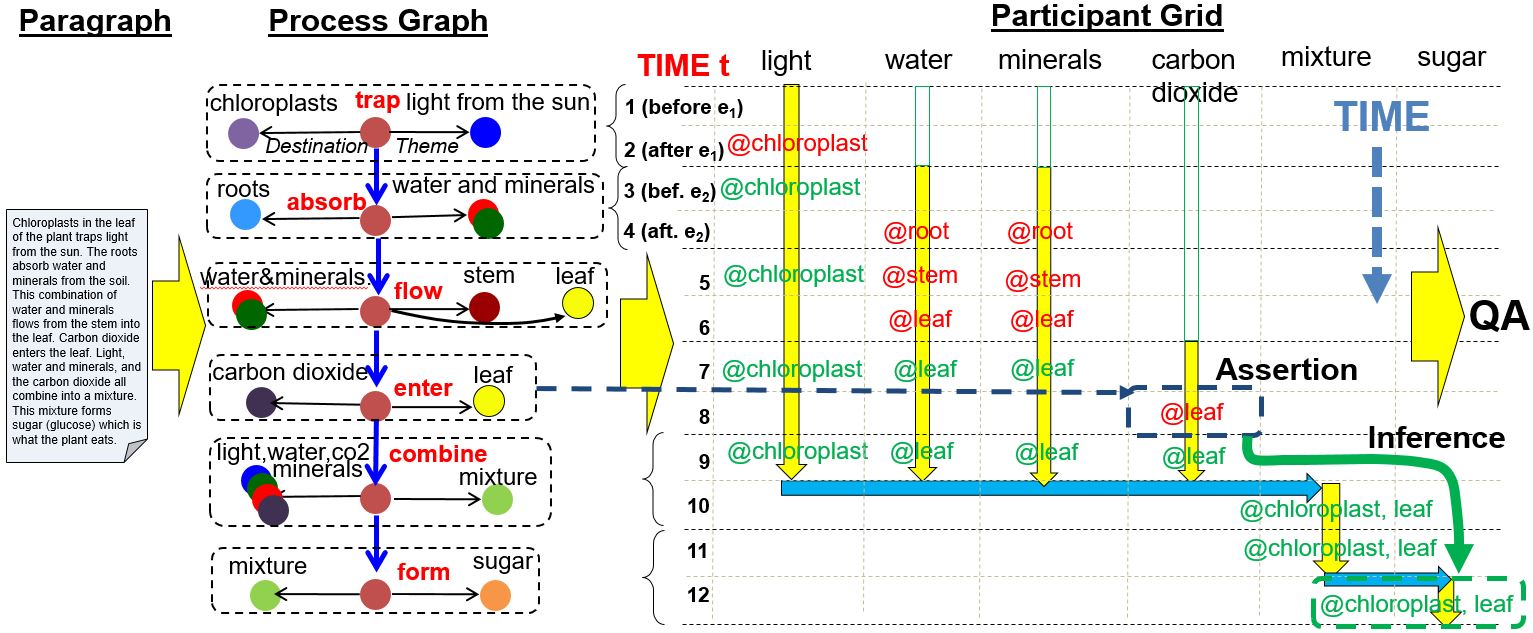}
\vspace{-3mm}
\caption{\ProComp~converts the paragraph to the Process Graph (same-colored nodes are coreferential) then to the Participant Grid,
displaying the states of the process before/after each event (time vertically downwards).
For brevity, @ denotes is-at(), arrows denote exists(), green denotes inferred literals. At line 8, the ``CO2 enters leaf'' 
step asserts that CO2 is therefore @leaf after the event. By inference, the sugar must therefore be produced at the leaf too (green box), a fact not explicitly stated in the text.}
\label{participant-grid}
\vspace{-4mm}
\end{figure*}

To perform this kind of reasoning, two types of knowledge are needed:
\begin{enu}
\item[(a)] what {\bf events} occur in the process (e.g., ``CO2 {\it enters} the leaf''), and
\item[(b)] what {\bf states} those events produced (e.g., ``CO2 {\it is at} the leaf'').
\end{enu}
Prior work on event extraction \cite{berant2014modeling,hogenboom2011overview,mcclosky2011event,Reschke2014EventEU}
addresses the former, allowing questions about event ordering to be answered.
Our work addresses the latter, by also representing the states that occur.
We do this with two key contributions:
\begin{enu}
\item[(1)] a VerbNet-derived rulebase, called the Semantic Lexicon, describing the preconditions
  and effects of different actions, expressed as linguistic patterns (Table~\ref{semantic-lexicon}).
\item[(2)] an illustration of how state modeling and machine reading
  can be integrated together, outperform RC methods that rely on surface cues alone.
\end{enu}
We evaluate our work on an early version of the ProPara dataset containing several hundred
paragraphs and questions about processes \cite{naacl}, and find that our system \ProComp~significantly outperforms two
strong reading comprehension (RC) baselines. 

Since this work was performed, we have developed neural systems that outperform
\ProComp, described elsewhere \cite{naacl}.
However, \ProComp~provides a strong baseline for other work,
the method illustrates how modeling and reading can be integrated
for improved question-answering, and the Semantic Lexicon remains a novel
and potentially useful resource. Its integration with neural systems remains a
currently unexplored opportunity for future work.


\section{Related Work}

General-purpose reading comprehension systems,
e.g., \cite{zhang2017exploring,Seo2016BidirectionalAF},
have become remarkably effective at factoid QA, driven by 
the creation of large-scale datasets, e.g., SQuAD \cite{rajpurkar2016squad},
TriviaQA \cite{JoshiTriviaQA2017}. However, they require extensive
training data, and can still struggle with queries requiring
complex inference \cite{hermann2015teaching}.
The extent to which these systems truly understand language remains 
unclear \cite{jia2017adversarial}.

\eat{
For the more specific task of understanding paragraphs about
processes, there have been several systems developed that 
explicitly extract some kind of process structure (events, arguments, and their relationships)
from the paragraph. ProRead \cite{berant2014modeling} converted process paragraphs
into a graph structure representing the process events and their relationships,
allowing questions about event ordering to be answered. It did not,
however, model how the state of the world changed during the process,
rendering some questions unanswerable.  
Similarly, \cite{kiddon2015mise} interpreted recipe paragraphs to produce 
an {\it action graph} of the recipe's actions, with the goal of
correctly interpreting the (often telegraphic and elided) recipe 
language, but intermediate states were not explicitly modeled.
Other projects too have focused on extracting domain-specific
event sequences from individual paragraphs, e.g., for biomedicine \cite{cohen2015darpa}, 
newswire articles \cite{caselli2017proceedings}, and ``how to'' guides
\cite{Chu2017DistillingTK}, or from large-scale corpora,
e.g., \cite{Chambers2008UnsupervisedLO}. 
Our goal is to similarly extract such sequences, but to 
then go further and reason about the states those
sequences imply, a challenging task because of the substantial
background event knowledge required. Early work in NLP had similar 
ambitions, e.g., \cite{schank2013scripts,dejong1979prediction}, 
but did not have access to the requisite commonsense knowledge,
limiting their generality.
}

More recently, several neural systems have been developed for
reading procedural text. 
Building on the general Memory Network architecture \cite{weston2014memory}
and gated recurrent models such as GRU~\cite{cho2014properties},
Recurrent Entity Networks (EntNet)~\cite{Henaff2016TrackingTW} uses a dynamic memory of hidden 
states (memory blocks) to maintain a representation of the world state, with a gated update
at each step. Similarly, Query Reduction Networks (QRN)~\cite{Seo2017QueryReductionNF} 
tracks state in a paragraph, represented as a hidden vector $h$.
Neural Process Networks (NPN)~\cite{bosselut2017simulating}, also models
each entity's state as a vector, and explicitly learns a neural model
of an action's effect from training data. NPN then computes the state change at each step from the
step's predicted action and affected entity(s), then updates the entity(s) vectors
accordingly, but does not model different effects on different entities by the same
action. Finally our own subsequent systems, \prolocal~and \proglobal,
learn neural models of the effects of actions from annotated training data \cite{naacl}.
In \cite{naacl}, we show that our rule-based system \ProComp~here also
outperforms EntNet and QRN, but not \prolocal~and \proglobal. The
integration of \ProComp's Semantic Lexicon with neural methods
remains an opportunity for future work.

For the background knowledge of how events change the world,
there are few broad-coverage resources available, although
there has been some smaller-scale work, e.g., \cite{Gao2016PhysicalCO} 
extracted observable effects from videos for verbs related to cooking.
One exception, though, is VerbNet \cite{schuler2005verbnet,kipper2008large}. 
The Frame Semantics part of VerbNet includes commonsense axioms about how events
affect the world, but it has had only limited use in NLP to date
e.g. \cite{schuler2000parameterized}. Here we show how this
resource can contribute the background knowledge required to
reason about change. 

Finally, there are numerous representational schemes developed for
modeling actions and change, e.g., STRIPS \cite{fikes1971strips,Lifschitz87onthe},
Situation Calculus \cite{levesque1998foundations}, and PDDL \cite{mcdermott1998pddl}.
Although these have typically been used for planning, here we use 
one of these (STRIPS) for language understanding and simulation,
due to its simplicity. 

\section{Approach}

We now describe how our system, ProComp (``process comprehension''), 
infers the states implied by text, and uses those to answer questions.
Figure~\ref{participant-grid} summarizes the approach for the running example in Figure~\ref{example}.

The input to ProComp is a process paragraph
and a question about the process, the output is the answer(s) to that question.
ProComp operates in three steps: (1) the sequence of events in the process is extracted from the paragraph
(2) ProComp creates a symbolic model of the {\it state} between each event, using the Semantic Lexicon,
a VerbNet-derived database showing how events change the world, and then performs inference over it (3) questions within ProComp's scope are 
mapped to computations over this model, and an answer is generated. The process states
are displayed as a {\it Participant Grid}, shown in Figure~\ref{participant-grid}.

\subsection{The Semantic Lexicon}

Before describing these steps, we describe the Semantic Lexicon itself.
The purpose of the Lexicon is to encode commonsense knowledge about the states that events produce.
A fragment is shown in Table~\ref{semantic-lexicon}.
An {\bf event} is an occurence that changes the world in some way, and 
a {\bf state} is a set of literals that describe the world (i.e., are true) at a particular time point.
The Semantic Lexicon represents the relationship between linguistic event descriptions and states using
using a STRIPS-style list of {\bf before} (preconditions)
and {\bf after} (effects) expressed as possibly negated literals
\cite{fikes1971strips}\footnote{
The {\bf after} list combines STRIPS' original ``add'' and ``delete'' lists by
using negation to denote predicates that become false (``deleted'').}.
While STRIPS used this knowledge for planning,
we use it for simulation: {\it Given} that an event occurs in the process, its 
{\bf before} conditions must have been true in the state beforehand, and its {\bf after} conditions after.

Each entry in the Lexicon consists of a (WordNet sense-tagged) {\bf verb} V, a {\bf syntactic pattern} of the form (S V O (Prep NP)*) showing 
how V might be used to describe an event, and the {\bf before} and {\bf after} conditions.
Most importantly, the syntactic pattern shows where the arguments of the before/after conditions may appear linguistically,
allowing syntactic elements of an event-describing sentence to be mapped to the arguments in the before/after literals.

\eat{
\begin{table}
\begin{tabular} {|l|l|l|l|} \hline
{\bf Predicate}		& {\bf Possible y Values} & \# entries \\ \hline 
is-at(x,y)		& {\it any NP} & 712 \\
exists(x)		& (no y argument) & 320 \\
size(x,y)		& increased; decreased & 90 \\
temperature(x,y)	& increased; decreased & 81 \\
phase(x,y)		& solid; liquid; gas & 65 \\ \hline
\end{tabular}
\caption{Five predicates are used to model changes in existence, location, temperature, size, and phase.}
 \label{predicates}
\vspace{-3mm}
\end{table}
}

The lexicon currently models change in existence, location, size, temperature, and phase (solid/liquid/gas).
It has entries for the preferred (most frequently used) senses of 2034 verbs\footnote{
These are all verbs that occurred in a collection of high-school-level tetbooks that we assembled,
with the most frequently used VerbNet sense assigned to each.}.
Note that some verbs may have multiple effects (e.g., ``melt'' affects both phase and temperature), while 
others may have none within the scope of our predicates (e.g., ``sleep'').

For existence and location, the Lexicon was first initialized using data 
from VerbNet, transforming its Frame Semantics entries for each verb.
VerbNet's assertions of the form $pred$(start(E),$a_1$,...,$a_n$) are converted to a {\bf before} assertion $pred$($a_1$,...,$a_n$),
when $pred$ is either exists() or location() (VerbNet does not explicitly model size, temperature, or phase changes).
Similarly, assertions with end(E) or result(E) arguments become {\bf after} assertions.
For example, for the pattern (Agent ``carve'' Product), VerbNet's Frame Semantics state that:

\vspace{2mm}
\centerline{\includegraphics[width=1.0\columnwidth]{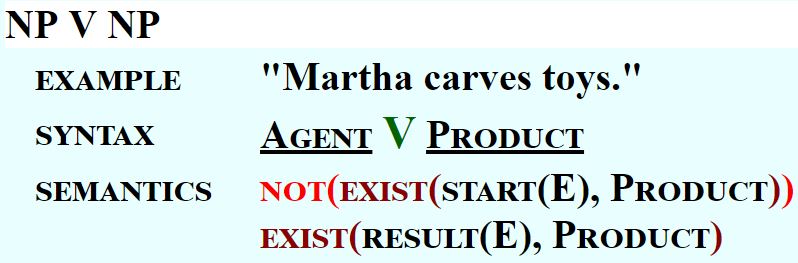}}
\vspace{1mm}

\noindent
We transform this to a rule:
\begin{quote}
{\bf IF} (Agent ``carve'' Product) \\
{\bf THEN before:} not exists(Product) \\
\hspace{1cm} {\bf \& after:} exists(Product)
\end{quote}


We then performed a manual annotation effort over several days to check and correct
those entries, and add entries for other verbs that affect existence and location.
At least two annotators checked each entry, and added new entries for verbs not annotated in VerbNet.
This was necessary as VerbNet is both incomplete and, in places, incorrect (its assumption
that all verbs in the same Levin Class \cite{levin1993english} 
have the same semantics only partially holds). Specifically, for the 2034 verbs in our Semantic Lexicon,
VerbNet only covered 891 of them (using 193 Levin classes). The manual annotation involved
checking and correcting these entries at the verb level, and added change axioms for
the remaining verbs as appropriate (note that many verbs have a ``no change'' entry,
if the verb's effects are outside those modelable using our predicates). 

For changes in temperature, size, and phase (not modeled in VerbNet's Frame Semantics), 
we first identified candidate verbs that might affect each through corpus analysis (collecting
verbs in the same sentence as temperature/size/phase adjectives), and then manually
created and verified lexicon entries for them ($\sim$20 hours of annotation effort).

While most verbs describe a specific change (e.g., ``melt''), there are a few 
general-purpose verbs (e.g., ``become'', ``change'', ``turn'', ``increase'', ``decrease'') 
whose effects are argument-specific. These are described using multiple patterns with type constraints 
on their arguments.

The final lexicon contains 4162 entries (including empty ``no change'' entries; note
that a single verb may have multiple entries for different subcategorizations).
It represents a substantial operationalization and expansion of (part of) VerbNet.

\subsection{Step 1. Process Extraction (the Process Graph)}

ProComp's first step is to extract the events in the paragraph, and assemble them 
into a Process Graph representing the process.
In ProComp a {\bf process} is a sequence of events,
represented as a {\bf process graph} $G$ = ($E$,$A$,$R_{ee}$,$R_{ea}$) where $E$ and $A$ are nodes
denoting events (here verbs) and verb arguments respectively, and $R_{ee}$ and $R_{ea}$ are edges
denoting event-event (e.g., next-event, depends-on) and event-argument (semantic role) relations
respectively.

ProComp first breaks the paragraph into clauses using ClausIE \cite{Corro2013ClausIECO}. 
Then, each clause is processed in two ways:
\begin{enumerate}
\item[(a)] OpenIE \cite{banko2007open} plus normalization rules convert each clause
into a syntactic tuple of the same (S V O PP*) form as in the Lexicon. The tuple is then
matched against patterns in the Lexicon to find the before/after assertions about
the described event.
\item[(b)] Semantic role labeling (SRL) is performed to identify the participants and their roles in the event that the clause describes. To increase the 
quality, we use an ensemble of SRL systems: neural network based DeepSRL \cite{Luheng2017DeepSRL}, 
linguistic feature based EasySRL \cite{Lewis2015JointAC}, and OpenIE \cite{Etzioni2011OpenIE}, 
with manually tuned heuristics to aggregate the signals together. 
Standard NLP techniques are used to normalize phrases, and a stop-list of 
abstract verbs is used to remove non-events. 
For example, in ``CO2 enters the leaf'', ``CO2'' is labeled as the Agent and
leaf as the Destination:
\begin{quote}
\centerline{\includegraphics[width=0.5\columnwidth]{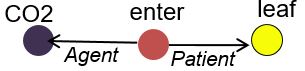}}
\end{quote}
\vspace{-4mm}
These roles are used to reason about the process (Step 2). 
\end{enumerate}
Finally, event (verb) nodes are connected by next-event links in the order they appeared in the text,
making the assumption that events will be presented in chronological order (non-chronological event ordering is 
out of scope of this work). Argument nodes with the same headword are merged (i.e., coreferential).
The full process graph $G$ for the earlier paragraph is shown in Figure~\ref{participant-grid}.

\subsection{Step 2. Simulation (the Participant Grid)} \label{method:representation-inference}

In Step 2, ProComp uses the process graph $G$ and Lexicon to infer the before and after states
of each event, and then reasons over these states, essentially simulating the process. To do this, for each event $e_i$ in $G$ that
matched an entry in the Lexicon, ProComp records facts true in the event's before/after states using 
$holds\mhyphen at(L,t)$ 
assertions in a state database, where $L$ is a before/after literal and $t$ is a time.
For event $e_i$, we define the before time $t=(2i\mhyphen 1)$ and the after time as $t=2i$.
For example, for ``carbon dioxide enters the leaf'' step $e_4$, ProComp finds and asserts
\begin{quote}
$holds\mhyphen at(is\mhyphen at(carbon~dioxide,leaf),8)$
\end{quote}
\noindent
i.e., after $e_4$ the carbon dioxide is {\it at} the leaf.

The contents of the database are displayed as a
{\it Participant Grid}, a 2D matrix with a column for each participant $p_j$ (verb argument), 
a row for each time step $t$ (time proceeding vertically downwards), and each cell (j,t) containing
the literals true of that participant $p_j$ at that time $t$ (i.e., where $p_j$ is the first
argument of the literal). 

For inference, 
second-order frame axioms project facts both forward and backwards in time:

\vspace{2mm} 
\noindent
$    \forall L,t~ holds\mhyphen at(L,t+1) \leftarrow \\
\hspace*{0.5cm} holds\mhyphen at(L,t) ~\&~ not ~{\sim}holds\mhyphen at(L,t+1) $ \\
$    \forall L,t~ holds\mhyphen at(L,t-1) \leftarrow \\
\hspace*{0.5cm} holds\mhyphen at(L,t) ~\&~ not ~{\sim}holds\mhyphen at(L,t-1) $
\vspace{2mm} 

\noindent
i.e., if literal $L$ holds at time $t$, and it is not inconsistent that $L$ holds at time $t+1$, then conclude 
that $L$ holds at $t+1$; similarly for $t-1$; where $not$ and $\sim$ denote negation as failure and 
strong negation respectively. Note that this formalism tolerates inconsistency 
in the process paragraph: if a projected fact clashes with what is known, the 
$holds\mhyphen at()$ assertions are not made and omitted from the intermediate Grid cells,
denoting lack of knowledge. 
Some examples of inferred facts are shown in green in Figure~\ref{participant-grid},
for example given CO2 is at the leaf at t=8, it is still at the leaf at t=9.

\eat{
\noindent
For example, given that  \\
\centerline{$holds\mhyphen at(is\mhyphen at(carbon~dioxide,leaf),8)$} \\ 
is true, ProComp infers for the next time step:
\centerline{$holds\mhyphen at(is\mhyphen at(carbon~dioxide,leaf),9)$} \\ 
i.e., the carbon dioxide stays at the leaf.
}

We also use seven rules expressing {\bf commonsense knowledge} to complete the Grid 
further, including using information from events that do not have explicit change effects.
These rules codify additional commonsense laws (the first 4 rules) and plausible inferences 
that follow from the pragmatics of discourse (the remaining 3 rules):
\begin{des}
\item[{\bf Location:}] If X is the Patient of event E, and E has initialLocation (resp. finalLocation)
L (even if E doesn't change a location), then X is at L before (resp. after) E.
\item[{\bf Existence:}] If X is an Agent/Patient of event E,
then (if no contrary info) X exists before and after E.
\item[{\bf Colocation:}] If X is converted to Y (i.e., X consumed + Y produced), 
and X is at L, then Y is at L (and vice versa). For example,
given that the CO2 is at the leaf at t=9, and is converted to a mixture, the mixture
will be at the leaf after the mixing at t=10 (collocation rule), Figure~\ref{participant-grid}.
\item[{\bf Creation:}] If only one event E involves participant X, and X is the Patient in E, then X is produced (created) at E.
\item[{\bf Destruction:}] If at least two events involves participant X, and X is the Patient in at least one of these, then 
                    X is consumed (destroyed) in the last of these events.
\item[{\bf Dependency:}] If Event $E_{i}$ is the first event involving participant X, and
     $E_{j}$ is the next event that has X as its Agent/Patient, then $E_{j}$ depends on $E_{i}$.
\item[{\bf Default Dependency:}] In the absence of other information, event $E_{i}$ depends on its previous event $E_{i-1}$.
\end{des}

\subsection{Step 3. Question Answering \label{qa}}

Given the participation grid, ProComp can answer 7 classes of question about the process: 
\begin{ite}
   \item What is produced/consumed/moved during this process? 
   \item Where is X produced/consumed? 
   \item Where does X move from/to?
   \item What increases/decreases in temperature? 
   \item What increases/decreases in size? 
   \item What changes from solid/liquid/gas to a solid/liquid/gas?
   \item What step(s) does step X depend on? 
\end{ite}

Questions are posed as instantiated templates (ProComp is designed for state modeling, and 
does not currently handle linguistic variability in questions).
Each template is twinned with a straightforward answer procedure that computes the answer from the participant grid.
For example, in Figure~\ref{participant-grid}, ``Where is sugar produced?'' is answered by the procedure for ``Where is X produced?''.
This procedure scans the X (``sugar'') column in Figure~\ref{participant-grid} to find where it comes into existence, and the location 
reported (chloroplasts,leaf). Each procedure thus codifies the semantics of each question class, i.e., maps the (templated) English 
to its meaning in terms of the predicates used to model the process. 

Although these technology components (VerbNet, event extraction, and process modeling) are familiar, 
this is the first time all three have been integrated together, allowing latent states in language
to be recovered and a new genre of questions to become answerable via modeling.

\section{Experiments}

\subsection{Dataset}

To evaluate this work, we used an earlier version of the ProPara dataset\footnote{available at \url{http://data.allenai.org/propara}} 
\cite{naacl}. ProPara is a new process dataset consisting of 488 crowdsourced
paragraphs plus questions about different processes.
The earlier version used here,
  called OldProPara, is similar but contains only 382 annotated paragraphs,
  uses a different train/dev/test split of 40/10/50, has additional
  anotations for size, temperature, phase, and event dependencies,
  and uses different (and easier) question templates (Section~\ref{qa}).
  OldProPara is available on request from the authors.
  
\subsection{Baselines}

We compared ProComp with two recent RC systems, BiDAF and ProRead. For subsequent
comparisons of ProComp with the neural systems EntNet and QRN, see \cite{naacl}.

\noindent
{\bf BiDAF} \cite{Seo2016BidirectionalAF} is a neural reading comprehension system
that is one of the top performers on the SQuAD (paragraph QA) dataset\footnote{
https://rajpurkar.github.io/SQuAD-explorer/}. We retrained BiDAF on our data, and found continued training 
(train on SQuAD then our data) produced the best results\footnote{
BiDAF F1 scores on Single Answer Questions were 62.2 (train on SQuAD then continue training on OldProPara, Table~\ref{results}), compared with
55.2 (retrain using only OldProPara), and 17.5 (original model, trained only on SQuAD). Similar differences hold on All Qns.},
so we report results with that configuration.

\noindent
{\bf ProRead} \cite{berant2014modeling} is a system explicitly designed for reasoning
about processes, in particular answering questions about event ordering and event arguments\footnote{
As ProRead only answers binary multiple choice (MC) questions, we converted each question 
into an N-way MC over all NPs in the passage, then used an all combinations binary tournament 
to find the winner \cite{landau1953dominance}.}.
We used ProRead's pretrained model, trained on their original data of
annotated process paragraphs, appropriate for our task. ProRead is not easily
extensible/retrainable as it requires extensive expert authoring of the entire 
process graph for each paragraph.

Because BiDAF and ProRead assume there is exactly one answer to a question
(while OldProPara questions can have 0, 1, or more answers), we report scores on 
both the entire test set, and also the subset with exactly 1 answer.

\begin{table}
{\small
\hspace*{-2mm}
\centering
\begin{tabular}{|l|lll|lll|} \hline
 & \multicolumn{3}{|c|}{Single Answer} & \multicolumn{3}{|c|}{} \\ 
 & \multicolumn{3}{|c|}{Questions} & \multicolumn{3}{|c|}{All Questions} \\ 
 & \multicolumn{3}{|c|}{(1675 questions)} & \multicolumn{3}{|c|}{(3562 questions)} \\ 
 & P & R & F1 & P & R & F1 \\ \hline
ProRead  & 33.9 & 33.9 & 33.9 & 17.8 &  16.8 & 17.3 \\
BiDAF   & 62.2 & 62.2 & 62.2 & 34.5 & 31.7 & 33.0 \\
{\bf ProComp} & 63.8 & 70.5 & {\bf 67.0} & 73.7 & 76.0 & {\bf 74.8} \\ \hline
\end{tabular}
}
\caption{Results for ProComp and two recent RC systems (ProRead and BiDAF) 
on the OldProPara data (test set). Scores are macroaveraged, F1 differences are statistically significant (p$<$0.05).}
\label{results}
\vspace{-3mm}
\end{table}

\subsection{Results}

The results are shown in Table~\ref{results}. ProComp significantly outperforms 
the baselines on both the full test set, and the single answer questions subset
(removing ProRead and BiDAF's disadvantage of producing just one answer). 
The low ProRead numbers reflect that it does not model states (it was 
primarily designed to reason about event arguments and event ordering), with
other questions, including most of the OldProPara questions, answered by
an Information Retrieval fallback \cite{berant2014modeling}. ProComp
also significantly outperforms BiDAF, but by a smaller amount. We analyze 
the respective stengths and weaknesses of ProComp and BiDAF in detail in the Analysis Section.

\subsection{Ablations}

We are also interested in two related questions:
\begin{enu}
\item[1.] How much have our extensions to the original VerbNet-derived lexicon helped?
\item[2.] How much has additional commonsense inference, beyond computing direct consequences of actions, helped?
\end{enu}
To evaluate this, we performed two (independent) ablations: 
(1) removing the extensions and corrections made to the original VerbNet axioms in the Lexicon
(2) disabling the inference rules about change; thus only states that directly follow from the Semantic Lexicon, rather 
than inferred via projection etc., are recovered.
Table~\ref{ablations} shows the results of these ablations. For the first ablation, the results 
indicate that the VerbNet extensions have significantly improved 
performance (+5.9\% F1 single answer questions, +2.4\% F1 all questions).
The second ablation similarly demonstrates additional inference using the commonsense rules 
improves performance (+6.6\% F1 single answer questions, +2.1\% F1 all questions). The relatively high
performance even without this additional inference suggests that many questions
in OldProPara ask about direct effects of actions. We analyze these further below.

\section{Analysis}

To understand the respective strengths of ProComp and BiDAF,
we analyzed answers on 100 randomly drawn questions. The relative performance on these is below:
\begin{center}
\begin{tabular}{l|cc} 
&              ProComp & ProComp \\
&               incorrect & correct \\ \hline
BiDAF incorrect &  21\% &  15\% \\
BiDAF correct &   14\% & 50\%  \\
\end{tabular}
\end{center}

\subsection{ProComp Successes}

There were 15 cases where ProComp was correct and BiDAF failed.
For the majority of these, BiDAF either did not recognize a verb's relation
to the question, or was distracted by other parts of the paragraph. For 
example (where {\bf T} is part of the paragraph, {\bf QA} is the question and the systems' answers,
and bold is the correct answer):

\begin{quote}\noindent\fbox{\parbox{0.9\columnwidth}{
{\bf T:} ...A {\bf roof} is built on top of the walls... \\
{\bf QA:} What is created during this process? {\bf roof} (ProComp), concrete (BiDAF)
}}\end{quote}
Here ProComp used the knowledge in the Lexicon that ``build'' is a creation event.
In 2 cases, ProComp performed more complex, multi-event reasoning, e.g.,:
\begin{quote}\noindent\fbox{\parbox{0.9\columnwidth}{
{\bf T:} ...transport the aluminum to a {\bf recycling facility}. 
The aluminum is melted down...
The melted aluminum is formed into large formations called ingots. The ingots are transported to another facility... \\
{\bf QA:} Where are the ingots moved from? {\bf a recycling facility} (ProComp), another facility (BiDAF)
}}\end{quote}
Here ProComp infers the aluminum is at the recycling facility (semantics of ``transport to'', first sentence), projects this to the formation of the ingots, and thus that the ingots are also at the recycling facility before they are moved.

{ \setlength{\tabcolsep}{4pt}
\begin{table}
{\small
\hspace*{-2mm}
\centering
\begin{tabular}{|l|lll|lll|} \hline
 & \multicolumn{3}{|c|}{Single Answer} & \multicolumn{3}{|c|}{} \\ 
\multicolumn{1}{|c|}{ProComp} & \multicolumn{3}{|c|}{Questions} & \multicolumn{3}{|c|}{All Questions} \\ 
\multicolumn{1}{|c|}{Configuration} & P & R & F1 & P & R & F1 \\ \hline
{\bf Full system}    & 63.8 & 70.5 & 67.0 & 73.7 & 76.0 & 74.8 \\
no VN extensions     & 58.2 & 64.3 & 61.1 & 71.2 & 73.6 & 72.4 \\
basic inference only & 59.4 & 61.4 & 60.4 & 72.8 & 72.6 & 72.7 \\ \hline
\end{tabular}
}
\caption{Results of two (separate) ablation experiments: 
(a) only use the original VerbNet axioms (ignore extensions/corrections)
(b) only use inferences from the Semantic Lexicon (ignore inference rules about projection, colocation, etc.).
F1 differences between the full and ablation versions are statistically significant (p$<$0.05).}
\label{ablations}
\vspace{-3mm}
\end{table}
}

\subsection{ProComp Errors}

\eat{
\begin{table}
\centering
\begin{tabular}{|l|l|} \hline
{\bf Cause of error:} & {\bf Extent} \\ \hline
{\bf NLP errors} & $\sim$20\% \\
\hspace{5mm}  Event detection, SRL failures & \\
{\bf Verb Semantics} & $\sim$30\% \\
\hspace{5mm} Lexicon rules incomplete/erroneous & \\
{\bf Complex Reasoning} & $\sim$40\% \\
\hspace{5mm} Reasoning outside ProComp's current scope & \\
{\bf Annotation and Scoring} & $\sim$10\% \\
\hspace{5mm}  Errors in the dataset, imperfect scoring metric & \\ \hline
\end{tabular}
\label{errors}
\caption{The four main classes of error in ProComp, and approximate size.}
\end{table}
}

There were also 35 cases where ProComp made mistakes. We identified four classes of error, 
described below. The impact percentages below were
judged using additional questions as well as the 35 failures.

\noindent
{\bf 1. NLP errors ($\sim$20\%):} Basic NLP errors can cause ProComp to fail, for example: 
\vspace{1mm} \\ \noindent\fbox{\parbox{0.9\columnwidth}{
{\bf T:} ...Rising {\bf air} cools... \\
{\bf QA:} What decreases in temperature? [no answer] (ProComp), {\bf air} (BiDAF)
}} \vspace{1mm} \\
For this question, OpenIE fails to produce a tuple for ``Rising air cools'', hence
no semantic implications of the event could be inferred. In general
the NLP pipeline can cause cascading errors; we later discuss how 
neural methods may be used to map directly from language to state changes, 
to reduce these problems.

\noindent
{\bf 2. Verb Semantics ($\sim$30\%):} In some cases there were subtle omissions in the 
Lexicon, for example:
\vspace{1mm} \\ \noindent\fbox{\parbox{0.9\columnwidth}{
{\bf T:} Water evaporates up to the {\bf sky}... \\
{\bf QA:} Where does the water move to? [no answer] (ProComp), {\bf sky} (BiDAF) 
}} \vspace{1mm} \\
Here ``evaporate'' had been annotated as a change of phase and temperature, but not
as a change of location. In contrast, BiDAF is likely using the surface cue ``to'' or ``up to'' in the text to answer correctly.

\noindent
{\bf 3. Complex knowledge and reasoning ($\sim$40\%):}

In some cases, knowledge and reasoning beyond ProComp is required to answer the question. For example: 
\vspace{1mm} \\ \noindent\fbox{\parbox{0.9\columnwidth}{
{\bf T:} ...The tiny ice particle...gathers more ice on the surface. Eventually the {\bf hailstone} falls... \\
{\bf QA:} What is created? supercooled water molecule (ProComp), updrafts ( BiDAF)
}} \vspace{1mm} \\
In this example, the conversion of the ice particle to hailstone is implicit, and hence ProComp does not recognize the hailstone's creation. 
Similarly:
\vspace{1mm} \\ \noindent\fbox{\parbox{0.9\columnwidth}{
{\bf T:} ...Fill the tray with {\bf water}....Place the tray in the freezer. Close the freezer door... \\
{\bf QA:} What decreases in temperature? [no answer] (ProComp), {\bf water} (BiDAF)
}} \vspace{1mm} \\
Reasoning that the water cools requires complex world knowledge (of freezers, containers, doors, and ramifications), beyond ProComp's abilities. In contrast, BiDAF guesses water - correctly, in this case, illustrating that sometimes surface cues can suffice.

\noindent
{\bf 4. Annotation and scoring errors ($\sim$10\%):}
In a few cases, the annotator-provided answers were questionable/incorrect (e.g., that tectonic  plates are ``consumed'' during an earthquake), or our headword-based scoring incorrectly penalized the systems for correct answers.

\section{Conclusion}

Our goal is to answer questions about change from paragraphs describing processes,
a challenging genre of text. We have shown how this can be done with two key contributions:
(1) a VerbNet-derived rulebase (the Semantic Lexicon), describing how events affect the world 
and (2) an integration of state-based reasoning with language processing, allowing
\ProComp~to infer the states that arise at each step. We have shown how this
outperforms two state-of-the-art systems that rely on surface cues alone.
The Semantic Lexicon is available from the authors on request.

Since this work was performed, we have subsequently developed two neural systems that outperform
\ProComp, described elsewhere \cite{naacl}. However, the Semantic Lexicon remains a novel
and potentially useful resource, \ProComp~provides a strong baseline for other work,
and the method illustrates how modeling and reading can be integrated
for improved question-answering. An integration of \ProComp's Semantic Lexicon with a
neural system remains a currently unexplored opportunity for further improvements
in machine reading about processes.

\bibliography{emnlp2017,references}
\bibliographystyle{emnlp_natbib}

\end{document}